\documentclass[10pt,twocolumn,letterpaper]{article}

\usepackage[pagenumbers]{arxiv} % To force page numbers, e.g. for an arXiv version

\usepackage{graphicx}
\usepackage{amsmath}
\usepackage{amssymb}
\usepackage{booktabs}

\usepackage{enumitem}
\usepackage{times}
\usepackage{epsfig}
\usepackage{amsfonts}
\usepackage{bm}
\usepackage{bbm}
\usepackage{multirow}
\usepackage{makecell}
\usepackage{ulem}
\usepackage{bbding}
\usepackage[table]{xcolor}

\usepackage{amssymb}
\usepackage{amsmath}
\usepackage{threeparttable}
\usepackage{multirow}
\usepackage{footnote}
\usepackage{eqnarray}
\usepackage{subcaption}
\usepackage{array}
\usepackage{makecell}
\usepackage{wasysym}
\newcommand{\ignore}[1]{}
\definecolor{cGreen}{RGB}{100,180,100}
\definecolor{cRed}{RGB}{220,50,0}

\definecolor{Klein_Blue}{rgb}{0.0, 0.129, 0.6}
\usepackage[pagebackref,breaklinks,colorlinks,bookmarks=false]{hyperref}
\hypersetup{
    colorlinks=true,
    linkcolor=magenta,
    urlcolor=magenta,
    citecolor=Klein_Blue
    }

\usepackage[capitalize]{cleveref}
\crefname{section}{Sec.}{Secs.}
\Crefname{section}{Section}{Sections}
\Crefname{table}{Table}{Tables}
\crefname{table}{Tab.}{Tabs.}

\begin{document}

\title{RTrack: Accelerating Convergence for Visual Object Tracking\\ via Pseudo-Boxes Exploration}

\author{
Guotian Zeng\textsuperscript{\rm 1},
    Bi Zeng\textsuperscript{\rm 1},
    Hong Zhang\textsuperscript{\rm 2},
    Jianqi Liu\textsuperscript{\rm 1},
    Qingmao Wei\textsuperscript{\rm 1},\\
    \textsuperscript{\rm 1}Guangdong University of Technology\\
    \textsuperscript{\rm 2}South University of Science and Technology\\
    zgt@mail2.gdut.edu.cn
}
%%\quad $^3$Peng Cheng Laboratory

\maketitle

%%%%%%%%% ABSTRACT
\begin{abstract}
  Single object tracking (SOT) heavily relies on the representation of the target object as a bounding box. 
  However, due to the potential deformation and rotation experienced by the tracked targets, the genuine bounding box fails to capture the appearance information explicitly and introduces cluttered background.
  This paper proposes RTrack, a novel object representation baseline tracker that utilizes a set of sample points to get a pseudo bounding box. RTrack automatically arranges these points to define the spatial extents and highlight local areas.
  Building upon the baseline, we conducted an in-depth exploration of the training potential and introduced a one-to-many leading assignment strategy.
  It is worth noting that our approach achieves competitive performance to the state-of-the-art trackers on the GOT-10k dataset while reducing training time to just 10\% of the previous state-of-the-art (SOTA) trackers' training costs. 
  The substantial reduction in training costs brings single-object tracking (SOT) closer to the object detection (OD) task. 
  Extensive experiments demonstrate that our proposed RTrack achieves SOTA results with faster convergence.
  %The code for RTrack is available at \href{https://github.com/relay/RepTrack}{\color{red}{https://github.com/relay/RepTrack}}.
\end{abstract}

\newcommand\blfootnote[1]{% 
\begingroup 
\renewcommand\thefootnote{}\footnote{#1}% 
\addtocounter{footnote}{-1}% 
\endgroup 
}
{
	% \noindent \blfootnote{\hspace{-6mm} 
	% $^\dagger$ Corresponding authors.
	% }
}

%%%%%%%%% BODY TEXT
\vspace{-2em}
\section{Introduction}
Visual object tracking (VOT) aims to locate and track  an arbitrary target over time in a video sequence, given only its initial appearance.
VOT potentially benefits the study of object detection, classification, and other related tasks such as action recognition and human-computer interaction, etc.

In single object tracking (SOT), the bounding box plays a crucial role in locating and tracking the object throughout a video sequence. It provides information about the target's position and scale in the current frame, and serves as input for feature extraction and classification. The widespread adoption of \textit{genuine} bounding box representation can be attributed to the following factors. Firstly, it aligns with commonly used performance metrics~\cite{GOT-10k,LaSOT,TrackingNet} that evaluate the overlap between estimated and ground truth boxes. Secondly, it offers convenience for feature extraction in deep networks~\cite{ResNet, vit, MAE} due to its regular orientation, allowing for easy subdivision of a rectangular window into a matrix of pooled cells.

\begin{figure}[!t]
    \centering
    \resizebox{\columnwidth}{!}{\includegraphics[width = \linewidth]{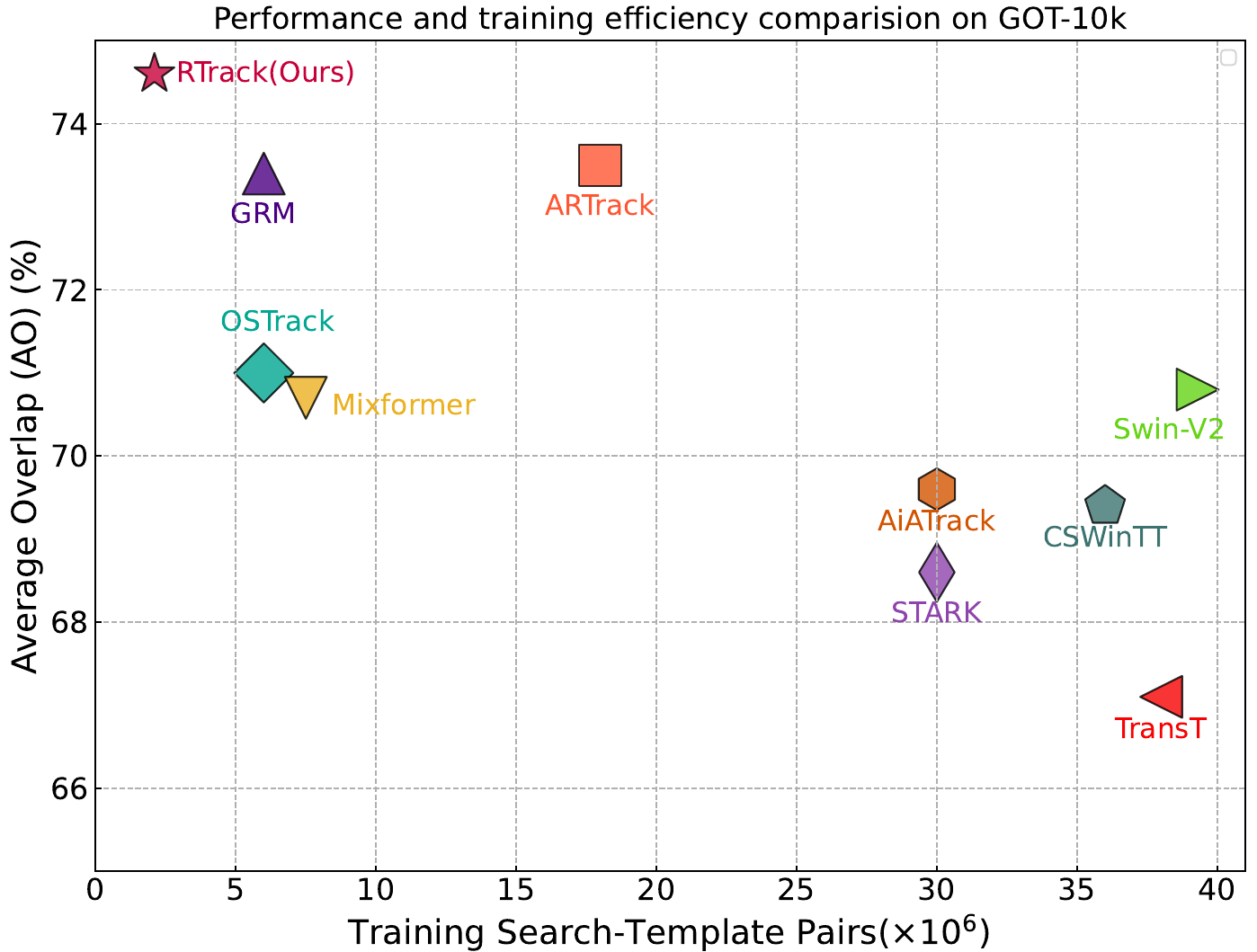}}
    \caption{Comparison of AO and
        training pairs of state-of-the-art trackers
        on GOT-10k under one-shot setting. Fewer Search-Template pairs means faster training convergence.
    }
    \label{fig::first_page_got10k}
\end{figure}
\begin{figure}[!t]
    \centering
    \resizebox{\columnwidth}{!}{\includegraphics[width = \linewidth]{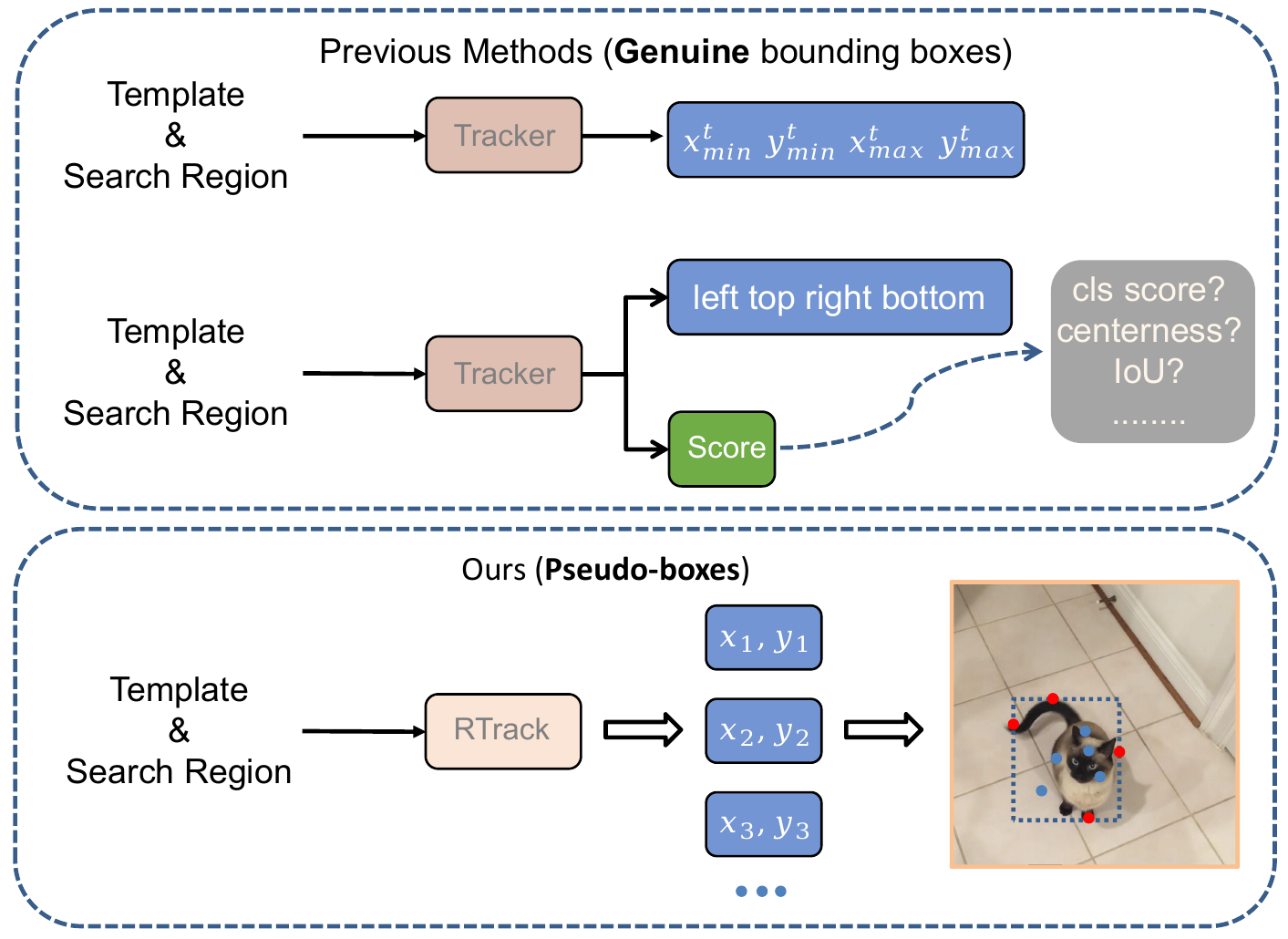}}
    \caption{
        Previous works on visual object tracking mainly relied on exploring the \textbf{genuine} bounding box. We propose to use a set of points to explore the appearance and ultimately generate \textbf{pseudo} bounding boxes. This is achieved through weak localization supervision from rectangular ground-truth boxes.}
    \label{fig::first_page_method_cmp}
\end{figure}

However, similar to object detection (OD), the bounding box representation only provides a rough localization of the target and does not account for its arbitrary appearance~\cite{RetinaNet, Mixformer,seqtrack}. When features are solely extracted from the bounding box, they may encompass background clutter, leading to a detrimental effect on the performance of single object tracking (SOT). Accurate prediction of bounding box plays a vital role in SOT, as the level of overlap between the estimated and ground truth bounding boxes is a widely adopted indicator.

Overall, the current tracker designs have the following problems: 1) The majority of current mainstream public visual object tracking benchmarks utilize rectangular bounding box annotations, as using point-based annotations can be costly and impractical. 2) The tracked objects have arbitrary orientations, and using genuine bounding boxes often introduce unnecessary background information. This requires the network to implicitly learn how to highlight the foreground, which slows down the convergence of training~\cite{TransT, STARK, Mixformer, aiatrack, seqtrack, artrack, sbt, videotrack}. 3) The improper distribution of positive and negative samples leads to training costs that are ten times higher than those of object detection algorithms~\cite{dino_mask,detr_co}.

In this paper, we propose \textit{RTrack}, a baseline tracker that utilizes a set of sample points to achieve accurate foreground localization. This design adaptively converts the sample points into \textit{pseudo} bounding boxes, allowing training using only bounding box annotations. Building upon our baseline, we further explore the potential of sample allocation strategies and propose a one-to-many leading assignment strategy that focuses on the trade-off between training efficiency and performance. Additionally, we implicitly incorporate correlation modeling into classification and regression. 
Our experiments demonstrate RTrack is effective and achieves state-of-the-art performance on several tracking benchmarks. 
Moreover, as depicted in \cref{fig::first_page_got10k}, compared to the recent state-of-the-art tracker Swin-V2~\cite{swinv2}, RTrack-256 exhibits significantly faster convergence, running 18.7 times faster, while achieving a superior AUC score of 3.8\% on the GOT-10k dataset. 
It is worth emphasizing that prior methods heavily rely on extensive training time to compensate for limited prior knowledge.
In contrast, our RTrack takes a different approach by thoroughly exploring the training potential and performance, as shown in ~\cref{fig::first_page_method_cmp}.

In summary, the contributions of this work are as follows:
\begin{itemize}
    \setlength{\itemsep}{0pt}
    \setlength{\parsep}{0pt}
    \setlength{\parskip}{0pt}
    \setlength{\parindent}{0pt}
    \item We propose \textit{RTrack}, a baseline tracker that leverages a set of sample points to achieve foreground localization. It adaptively converts the sample points into pseudo bounding boxes, allowing training using only bounding box annotations.
    \item We present a progressive one-to-many leading assignment strategy that effectively discriminates between positive and negative samples. Additionally, we implicitly incorporate correlation modeling into classification and regression.
    \item Comprehensive experiments demonstrate that RTrack achieves state-of-the-art (SOTA) performance with a remarkably fast training convergence, such as 35 epochs (0.25 RTX3090 GPU days) on the GOT10k dataset. 
\end{itemize}

\begin{figure*}[tb]
    \centering
    \includegraphics[width =0.95 \linewidth]{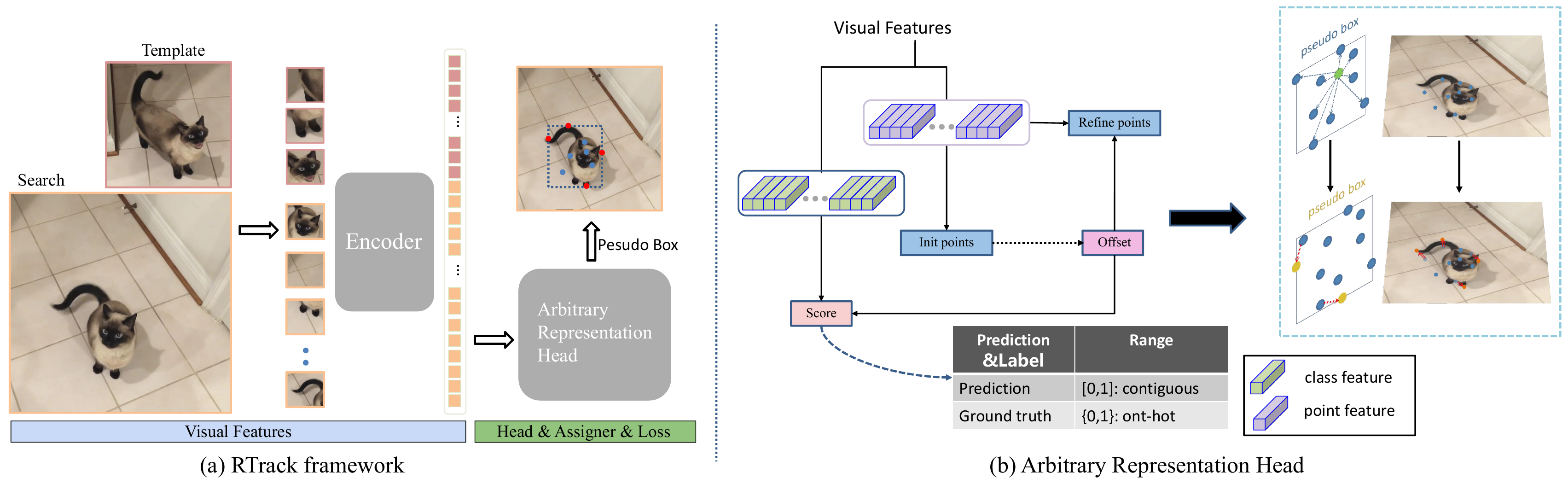}
    \caption{
        The architecture of the proposed \textbf{RTrack} consists of two key components: the Encoder and the subsequent arbitrary representation head. The Encoder extracts visual features. The two-stage arbitrary representation head generates pseudo boxes for the tracked objects, which are used for subsequent sample assigner and loss computation.
    }
    \label{fig::framework}
\end{figure*}

\section{Related Work}
\label{sec::related-work}

\subsection{Non-axis aligned representations}

Traditional methods~\cite{RetinaNet, SSD} primarily focus on detecting axis-aligned objects. However, these methods may encounter difficulties in detecting non-axis-aligned targets that are densely distributed in complex backgrounds.
More recently, there has been a focus on bottom-up approaches in object detection, including methods like CornerNet~\cite{CornerNet,cswintt, mat}, ExtremeNet~\cite{ExtremeNet}, and CenterNet~\cite{CenterNet, OSTrack, GRM}.
CornerNet predicts the top-left and bottom-right heatmaps of objects to generate the corresponding bounding boxes. However, it's important to note that these corner points still essentially represent a rectangular bounding box.
ExtremeNet aims to locate the extreme points of objects in both horizontal and vertical directions, with supervision from ground-truth mask annotations. This approach focuses on identifying specific points on the object's boundary, which allows for potentially more precise localization.

\subsection{Pseudo boxes generation without annotations}
Bottom-up detectors offer certain advantages such as a reduced hypothesis space and the potential for more precise localization. However, they often rely on manual clustering or post-processing steps to reconstruct complete object representations. In contrast, RepPoints~\cite{reppointv1, reppointsv2, ma2020rpt} provides a flexible object repre	sentation without the need for handcrafted clustering. RepPoints can learn extreme points and key semantic points automatically, even without additional supervision beyond ground-truth bounding boxes. 
In our proposed RTrack, we integrate RepPoints with deformable convolution~\cite{DCNv1, DCNv2}, aligning with the point representation and effectively aggregating information from multiple sample points.  Moreover, RepPoints can easily generate the ``pseudo box", making it naturally compatible with existing SOT benchmarks.

\subsection{Label assignment}
Many detection methods commonly utilize a hand-crafted threshold~\cite{RetinaNet, TransT} for selecting positive samples.
However, hand-crafted settings do not guarantee the overall quality of training samples, especially in the presence of noise and hard cases~\cite{cvpr2020learningfromnoisy, macvpr2021iqdet}. Hard samples, which often have a low IoU with anchors or cover a limited number of feature points, lead to a scarcity of positive samples. ATSS~\cite{ATSS_CVPR_2020}, Autoassign~\cite{autoassign_2020_arxiv}, and OTA~\cite{ota_2021_cvpr} have emphasized the importance of label assignment for improving detector performance. These methods employ an optimization strategy to select high-quality samples.
In single object tracking (SOT), selecting high-quality samples becomes crucial due to the diverse orientations and sparse distribution of objects.
In this paper, we propose an effective one-to-many leading samples assignment scheme that is specifically designed for our proposed baseline RTrack. Our scheme aims to select positive samples that accurately represent the target's quality and facilitate faster convergence.

\section{Methodology}

This section provides a detailed presentation of our proposed RTrack, which is divided into three parts:
1) Baseline tracking framework: we present a novel object representation baseline tracking framework, which serves as the foundation for the following explorations;
2) Building upon the baseline tracker, we further propose a progressive one-to-many leading assignment strategy to enhance performance and convergence speed. Additionally, we introduce an approach for modeling the correlation between classification and regression tasks to implicitly encode mutual affinity;
3) Finally, we describe the training loss of RTrack.
For more details about adaptive object representation, please refer to the appendix.

\subsection{Baseline tracker}
\label{Sec:pbt}
Our RTrack consists of two key components: a simple Encoder for simultaneous feature extraction and
relation modeling, and an arbitrary representation head designed to generate pseudo target bounding boxes. The overview of the model is shown in~\cref{fig::framework}.

\subsubsection{Encoder}
Our RTrack employs a progressive multi-layer architecture design~\cite{vit, OSTrack, simTrack}, where each layer operates on the same-scaled feature maps with the same number of channels.
Given the initial templates of size $H_t$$\times$$W_t$$\times3$ and the search region of size $H_s$$\times$$W_s$$\times3$, we first map them to non-overlapping patch embeddings using a convolutional operation with stride 16 and kernel size 16. This convolutional token embedding layer is introduced in each input to increase the number of channels while reducing the spatial resolution. Next, we flatten the embeddings and concatenate them to produce a fused token sequence of size $L$~$\times$~$C$, where $L$ is the fused sequence length and $C$ is the number of channels.
Finally, we obtain a deep interaction-aware feature of size $L'$ $\times$ C, where $L'$ is less than $L$. Before inputting the prediction head, the search region feature is recovered, split, and reshaped to a 2D spatial feature map of size $\frac{H_s}{16}~$\texttimes$~\frac{W_s}{16}$.

\begin{figure*}[htbp]
    \centering
    \includegraphics[width = \linewidth]{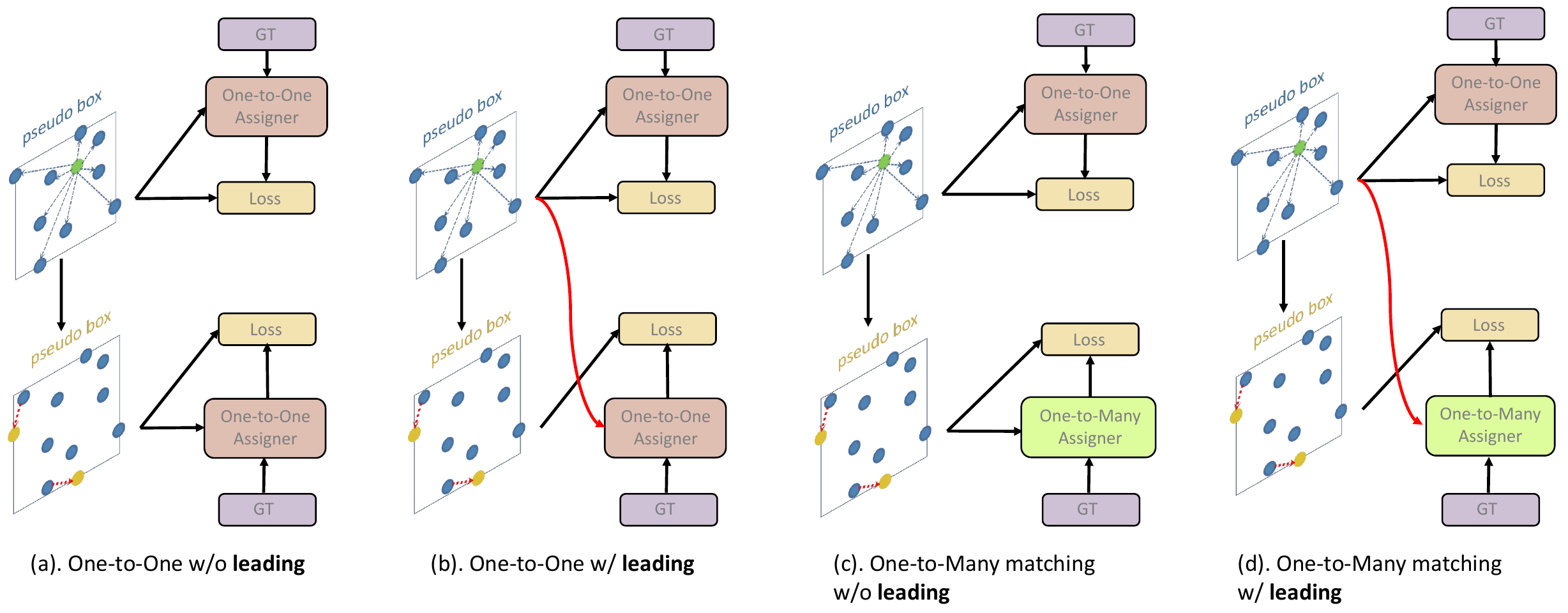}
    \caption{
        The sample assignment strategies of the init stage and refine stage.
        Compared with the assigner strategy (a), the schema in (b) introduces our proposed \textbf{leading strategy}, which leverages the strong learning ability of the init stage. In the init stage, we consistently maintain the center point-based initial representation and assign a \textbf{unique} positive sample, which we refer to as the \textbf{one-to-one} assigner strategy. In the refine stage (c), (d), we propose the \textbf{one-to-many} assigner strategy to accelerate the convergence speed.}
    \label{Fig::assigner}
\end{figure*}

\subsubsection{Arbitrary Representation Head}
As illustrated in~\cref{fig::framework} (b), our arbitrary representation head consists of two stages:
1) \textit{Init stage}: generating the first-points-set by refining from the object center point hypothesis. 
For the init stage, we use center points as the initial representation of objects
\footnote{A feature map bin is considered \textit{positive} if the center point of the ground-truth object falls within this feature map bin.}.
Starting from the center point, the first set of \textit{TrackPoints} is obtained by regressing offsets over the center point.
The head automatically arranges \textit{TrackPoints} to define the foreground of the target object. These sample points are then transformed to generate a pseudo box, which serves as the initial pseudo bounding box.
2) \textit{Refine stage:} generating the second-points-set by refining from the first-points-set.
For the second localization stage, the head further refines the representation by adapting the bounding box to better fit the target, and we adopt the approach used in common trackers, where the feature bin with the highest score is considered as the unique \textit{positive} sample.
This multi-stage design allows for progressive improvement and refinement of the object representation.

\begin{figure}[!t]
    \centering
    \resizebox{\columnwidth}{!}{\includegraphics[width = \linewidth]{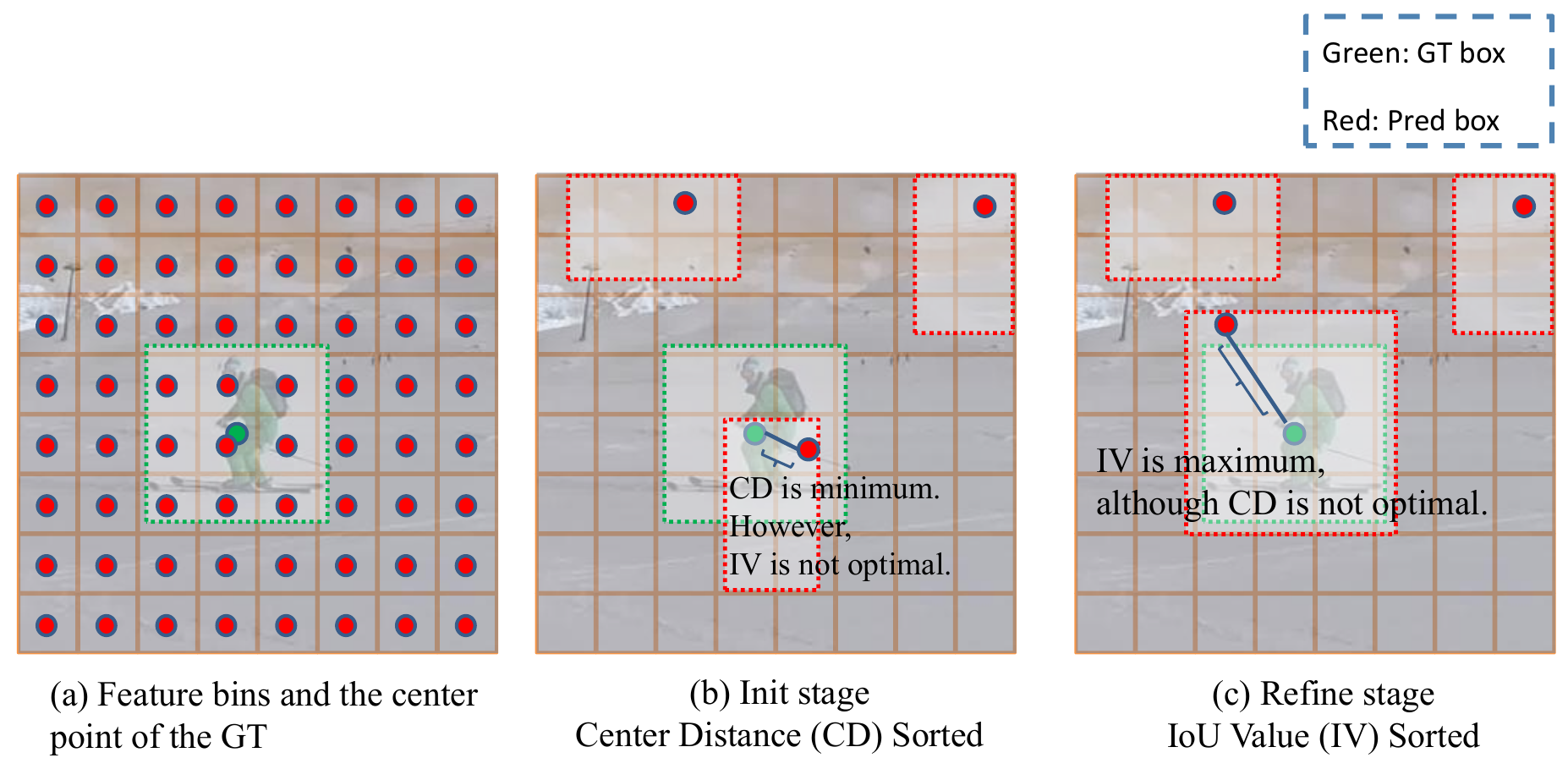}}
    \caption{The relationship between feature bins and the ground truth (GT) center is as follows: In the init stage, we select the feature bin with the smallest center distance (CD) as the unique positive sample. In the refine stage, we select multiple feature bins with the smallest CD or IoU value (IV) according to the one-to-many strategy and assign them as positive samples.}
    \label{Fig::bins}
\end{figure}

\subsubsection{Discussion of the proposed baseline tracker }
\begin{figure}[!t]
    \centering
    \resizebox{\columnwidth}{!}{
        \includegraphics[width = \linewidth]{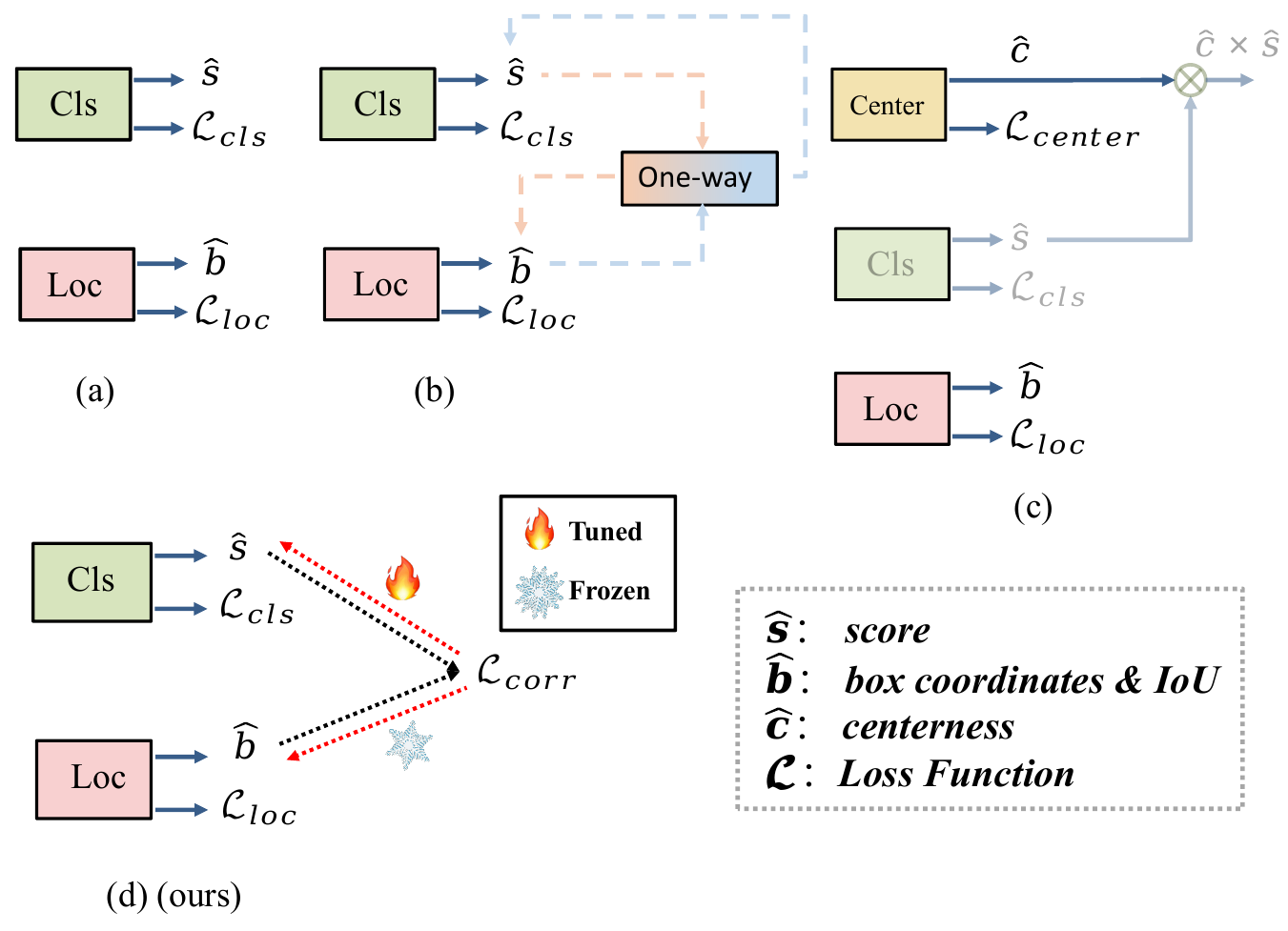}
    }
    \caption{The relationship between classification (\textbf{cls}) and localization (\textbf{loc}) tasks can be approached in several ways: (a) Independent optimization of both tasks; (b) One-way information flow; (c) Additional branches; (d) Our mutual affinity design.}
    \label{Fig::corr}
\end{figure}

1) Sample allocation strategy: our proposed baseline tracker employs a \textbf{one-to-one} matching strategy for allocating positive and negative samples in two stages, as shown in \cref{Fig::assigner} (a). This approach results in each stage having only one positive sample, which leads to a significant discrepancy in the training costs between single object tracking (SOT) and efficient object detection (OD) algorithms\cite{yolov7,detr_co,dino_mask}. To tackle this issue, we propose an \textbf{one-to-many leading} allocation strategy that is specifically designed to unleash the potential of our baseline tracker in the following section;
2) In our proposed baseline tracker, we only allow the flow of localization information (IoU) to the classification branch (score) in a one-way manner, as illustrated in ~\cref{Fig::corr} (b). This information imbalance can limit the performance of the tracker. We will introduce an implicit approach to model the \textbf{mutual affinity} between these two branches, addressing the correlation between classification (score) and localization (IoU) in the following section.

\subsection{One-to-many leading assigner strategy}
\label{sec::framework}
The current mainstream trackers commonly employ a one-to-one matching strategy that allows for a unique positive sample. However, it leads to a sparse distribution of samples and significantly increases the training costs required for these trackers.
As shown in \cref{Fig::assigner} (a), our baseline tracker separates two stages (an init stage and a refine stage), and then use their own prediction results and the ground truth to execute one-to-one label assignment.
Based on the observations above, we propose a progressive one-to-many leading assignment strategy, as shown in \cref{Fig::assigner} (d):
\begin{enumerate}
    \setlength{\itemsep}{0pt}
    \setlength{\parsep}{0pt}
    \setlength{\parskip}{0pt}
    \setlength{\parindent}{0pt}
    \item In the init stage, we select the feature bin closest to the ground truth object's center point based on center distance (CD) as the unique positive sample, termed as \textbf{one-to-one} assigner strategy, as illustrated in \cref{Fig::bins} (b).
    \item In the refine stage, we adopt an \textbf{one-to-many} assigner strategy by allowing more feature bins to be treated as positive targets by relaxing the constraints of the training potentials. We select the top K candidate positive samples with the highest intersection over union (IoU) values between the predicted bounding boxes and the ground truth box, as depicted in \cref{Fig::bins} (c). We then set the \textit{dynamic} threshold as the sum of the mean and variance of the candidate samples.
\end{enumerate}

We note that the additional positive samples may produce bad prior at the final prediction. Therefore, in order to make those extra coarse positive grids have less impact, we put restrictions in the extra coarse positive samples by following ``\textbf{leading}''.

The leading label assigner, depicted in \cref{Fig::assigner} (d), utilizes the predictions from the init stage and the ground truth information to calculate the dynamic labels. These labels are generated through an optimization process, leveraging the guidance from the init stage to inform the refine stage.
Specifically, the predictions from the init stage are used as guidance to generate hierarchical labels, which are then utilized in the refine stage.
The reason to do this is that the init stage has a relatively strong learning capability, so the label generated from the init stage should be more representative of the distribution and correlation between the source data and the target. By letting the refine stage directly learn the information that the init stage has learned, the init stage will be more able to focus on information that has not yet been learned.

\subsection{Mutual affinity between classification and localization}
\label{Sec::corr}
Instead of training classification and localization tasks independently (\cref{Fig::corr} (a)), most trackers~\cite{SiamFC, sparsett} optimize a weighted sum of classification and localization losses during training. Recent works~\cite{UAST, tatrack, GRM} suggest that  performance improves when these two loss functions are forced to interact with each other as illustrated in \cref{Fig::corr} (b).
And several methods (\cref{Fig::corr} (c)) have been recently proposed to improve the correlation between these tasks: training an auxiliary head to regress the localization qualities of the positive examples, e.g. centerness, or IoU, has proven useful \cite{IoUNet, FCOS, ATSS}.
Indeed, there are approaches ~\cite{Mixformer, STARK,aiatrack} that incorporate plug-in components~\cite{precise_roi_pooling} and train an additional branch for predicting IoU. While these two-stage training approaches can potentially improve performance, they also increase training costs and model complexity. They may not fully capture the correlation between these two branches, which can limit their performance.

As shown in ~\cref{Fig::corr} (d), given loss function $\mathcal{L}$, correlation loss ($\mathcal{L}_{corr}$) is simply added using a weighting hyper-parameter $\lambda_{corr}$: 
\begin{equation}
\label{eq:LossCorr}
\mathcal{L}^{'} = \mathcal{L} + \lambda_{corr} \mathcal{L}_{corr}.
\end{equation}
$\mathcal{L}_{corr}$ is the correlation loss defined as: 
\begin{equation}
\label{eq:LossCorr2}
\mathcal{L}_{corr} = 1 - \rho(\hat{\mathrm{IoU}}, \hat{\mathrm{s}}),
\end{equation}
where $\hat{\mathrm{s}}$ is the confidence score of the foreground, $\hat{\mathrm{IoU}}$ is IoU of the predicted pseudo boxes.
And $\rho(\cdot, \cdot)$ is a correlation coefficient~\cite{corr_loss} pertaining to the positive sample, defined as follow:
\begin{small} 
    \begin{eqnarray}
    \nonumber
    & & \bar{s} = \frac{1}{|\mathcal{C}|} \sum_{c \in \mathcal{C}}{\hat{s}^{c}}  \qquad \bar{b} = \frac{1}{|\mathcal{C}|} \sum_{c \in \mathcal{C}}{\hat{b}^{c}}\\
    \nonumber
    & & v_{s} = \hat{s}-\bar{s}  \qquad v_{b} = \hat{b}-\bar{b} \\
    %& & v_{s} = \hat{s}-\bar{s} \qquad v_{b} = \widehat{b}-\bar{b}\\
    \rho(\widehat{s},\widehat{b}) & = & \frac{\frac{2 \times \sum_{c \in \mathcal{C}}{(v_{s}^{c} \times v_{b}^{c})}}{\sqrt{\sum_{c \in \mathcal{C}}v_{s}^{c^2}} \times
            \sqrt{\sum_{c \in \mathcal{C}}v_{b}^{c^2}}} \times
        \operatorname{std}(\widehat{s}) \times \operatorname{std}(\widehat{b})}{\operatorname{var}(\hat{s})+\operatorname{var}(\widehat{b})+(\bar{s}-\bar{b})^{2}} 
    \end{eqnarray} 
\end{small}
We only use the gradients of $\mathcal{L}_{corr}$ wrt. classification score, i.e., we backpropagate the gradients through only the classification branch, termed as \textit{gradient truncation}, as shown in \cref{Fig::corr} (d).

\begin{table*}[!ht]
  \centering
  \small
  \resizebox{\linewidth}{!}{
      \begin{tabular}{c||cccccccc||ccc}
          \toprule
          ~ & 
          \multirowcell{2}[0pt][c]{TransT\\~\cite{TransT}} & 
          \multirowcell{2}[0pt][c]{STARK\\~\cite{STARK}} &
          \multirowcell{2}[0pt][c]{SBT\\~\cite{sbt}} & 
          \multirowcell{2}[0pt][c]{OSTrack\\~\cite{OSTrack}} & 
          \multirowcell{2}[0pt][c]{VideoTrack\\~\cite{videotrack}} & 
          \multirowcell{2}[0pt][c]{TATrack\\~\cite{tatrack}} & 
          \multirowcell{2}[0pt][c]{SwinV2\\~\cite{swinv2}} & 
          \multirowcell{2}[0pt][c]{ARTrack\\~\cite{artrack}} & 
          \multirowcell{2}[0pt][c]{RTrack\\-256} & 
          \multirowcell{2}[0pt][c]{RTrack\\-384} \\
          ~ & ~ & ~ & ~ & ~ & ~ & ~ & ~ & ~ & ~ \\
          \midrule
          Epochs & 1000 & 500 & 600 & 100 & 300 & 60 & 300 & 300 & 35 &  60\\ 
          Pairs / epoch($\times10^{4}$) & 3.8 & 6 & 5 & 6 & 6 & 30 & 13.1 & 6 & 6 & 6 \\ 
          Training Pairs($\times10^{6}$)~$\downarrow$ & 38 & 30 & 30 & 6 & 18 & 18 & 39.3 & 18 & \textbf{2.1} & \textbf{3.6} \\ 
          \midrule
          mAO(\%)~$\uparrow$ & 67.1 & 68.8 & 69.9 & 71.0 & 72.9 & 73.0 & 70.8 & 73.5 & \textbf{74.6} & \textbf{76.4} \\ 
          SR$_{0.5}$(\%)~$\uparrow$ & 76.8 & 78.1 & 80.4 & 80.4 & 81.9 & 83.3 & - & 82.2 & \textbf{84.4} & \textbf{86.0} \\ 
          SR$_{0.75}$(\%)~$\uparrow$ & 60.9 & 64.1 & 63.6 & 68.2 & 69.8 & 68.5 & - & 70.9 & \textbf{72.1} & \textbf{74.1}\\ 
          \bottomrule
      \end{tabular}
  }
  \caption{Comparision with state-of-the-art trackers on the GOT-10k test set. 
      All the results reported strictly adhere to the one-shot protocol, which means that our models were only trained using the GOT-10k training set and no additional data was used for training. }
  \label{Tab::sota_got10k}
\end{table*}

\subsection{Training loss}
In this section, we describe the training procedure for our proposed RTrack.
First, we feed the output $\frac{H_s}{16}~$\texttimes$~\frac{W_s}{16}$~\texttimes$~C$ of the backbone network into our proposed arbitrary representation head.
We adopt the weighted focal loss~\cite{focal} for classification as follows:
\begin{equation}
    \resizebox{\hsize}{!}{
        $
        \mathcal{L}_{cls}=-\sum_{x y}\left\{\begin{array}{ll}
        \left(1-\boldsymbol{P}_{x y}\right)^{\alpha} \log \left(\boldsymbol{P}_{x y}\right), & \text { if } \hat{\boldsymbol{P}}_{x y}=1 \\
        \left(1-\hat{\boldsymbol{P}}_{x y}\right)^{\beta}\left(\boldsymbol{P}_{x y}\right)^{\alpha} \log \left(1-\boldsymbol{P}_{x y}\right), & \text { otherwise }
        \end{array}\right.
        $
    }
\end{equation}
\noindent where $\alpha=2$ and $\beta=4$  are the regularization parameters in our experiments
as in \cite{OSTrack}. 

With the predicted bounding box, the generalized IoU loss~\cite{generalized_iou} are employed for bounding box regression for fair comparison as follows:
\begin{equation}
\mathcal{L}_{\text {det }}=\lambda_{init} \mathcal{L}_{\text {init }}+\lambda_{refine} \mathcal{L}_{refine}
\end{equation}

\noindent where $\lambda_{init}=1$ and $\lambda_{refine}=2$ as in \cite{OSTrack}.

Finally, we model the correlation between classification $\mathcal{L}_{\text {cls}}$ and localization  $\mathcal{L}_{\text {det}}$.
The full loss function is:
\begin{equation}
\mathcal{L}_{\text {all }}=\mathrm{\lambda}_{\mathrm{cls}}\mathcal{L}_{\mathrm{cls}}+\mathrm{\lambda}_{\mathrm{det}}\mathcal{L}_{\text {det }}+\mathrm{\lambda}_{\mathrm{corr}} \mathcal{L}_{corr}
\end{equation}

\noindent where $\lambda_{cls} = 2$, $\lambda_{det} = 1$ and $\lambda_{corr} = 0.5$.

\begin{table*}[!ht]
  \centering
  \small
  \resizebox{\linewidth}{!}{
      \begin{tabular}{c||cccccccc||cccc}
          \toprule
          ~ & \multirowcell{3}[0pt][c]{TransT\\~\cite{TransT}} & 
          \multirowcell{3}[0pt][c]{CS\\WinTT\\~\cite{cswintt}} & 
          \multirowcell{3}[0pt][c]{Mix\\Former\\~\cite{Mixformer}} &
          \multirowcell{3}[0pt][c]{Sim\\Track\\~\cite{simTrack}} & 
          \multirowcell{3}[0pt][c]{OS\\Track\\~\cite{OSTrack}} & 
          \multirowcell{3}[0pt][c]{MAT\\~\cite{mat}} & 
          \multirowcell{3}[0pt][c]{Swin\\V2\\~\cite{swinv2}} & 
          \multirowcell{3}[0pt][c]{Seq\\Track\\~\cite{seqtrack}} & 
          \multirowcell{3}[0pt][c]{RTrack\\-256} & 
          \multirowcell{3}[0pt][c]{RTrack\\-256} & 
          \multirowcell{3}[0pt][c]{RTrack\\-384}\\
          ~ & ~ & ~ & ~ & ~ & ~ & ~ & ~ & ~  & ~ & ~ & ~ \\ 
          ~ & ~ & ~ & ~ & ~ & ~ & ~ & ~ & ~  & ~ & ~ & ~ \\
          \midrule
          Epochs~ & 1000 & 600 & 500 & 500 & 300 & 500 & 300 & 500 & 100 & 300 & 300\\ 
          Pairs / epoch($\times10^{4}$) & 3.8 & 6 & 6 & 6 & 6 & 6.4 & 13.1 & 3 & 6 & 6 & 6\\
          Training Pairs($\times10^{6}$)~$\downarrow$ & 38 & 36 & 30 & 30 & 18 & 32 & 39.3 & 15 & \textbf{6} & 18 & 18\\ 
          \midrule
          AUC(\%)~$\uparrow$ & 81.4 & 81.9 & 83.1 & 82.3 & 83.1 & 81.9 & 82.0 & 83.3 & 83.7 & \textbf{84.2} &  \textbf{85.0}\\ 
          PRE$_{norm}$(\%)~$\uparrow$ & 86.7 & 86.7 & 88.1 & 86.5 & 87.8  & 86.8 & - &  88.3 & 88.6 & \textbf{88.8} & \textbf{89.4}\\ 
          PRE(\%)~$\uparrow$ & 80.3 & 79.5 & 81.6 & - & 82.0  & - & - &  82.2 & 82.7 & \textbf{83.2} & \textbf{85.0} \\ 
          \bottomrule
      \end{tabular}
  }
  \caption{Comparision with state-of-the-art trackers on the TrackingNet test set.}
  \label{Tab::sota_tknet}
\end{table*}

\begin{table*}[!ht]
  \centering
%	\resizebox{\textwidth}{!}{
      \begin{subtable}[t]{\linewidth}
          \centering
          \begin{tabular}{c||cccc||ccc|cc|cc||c}
              \toprule
              ~ & ~ & ~ & \multirow{3}{*}{\makecell[c]{Leading \\ Assigner \\ (IV)}} & ~ & \multicolumn{3}{c|}{GOT-10k} & \multicolumn{2}{c|}{LASOT}  & \multicolumn{2}{c||}{OTB100} & ~ \\ 
              \multirow{-2}{*}{\#} & \multirow{-2}{*}{\makecell[c]{baseline$\ddagger$}} & \multirow{-2}{*}{\makecell[c]{MaxIoU \\ Assigner}} & ~ & \multirow{-2}{*}{\makecell[c]{Mutual \\ Affinity\tnote{2}}} & \makecell[c]{mAO \\ (\%)} & \makecell[c]{mSR$_{50}$ \\ (\%)} & \makecell[c]{mSR$_{75}$ \\ (\%)} & \makecell[c]{SUC \\ (\%)} & \makecell[c]{PRE \\ (\%)} & \makecell[c]{SUC \\ (\%)} & \makecell[c]{RPE \\ (\%)} & \multirow{-2}{*}{\makecell[c]{$\Delta$}} \\ 
              \midrule
              1 & \checkmark & - & - & - & 70.9  & 80.1 & 68.3 & 65.5 & 69.6 & 65.9 & 85.4 & -  \\  
              2 & \checkmark & - & - & \checkmark & 68.8 & 77.8 & 65.7 & 64.9 & 68.2 & 65.9 & 85.4 & \textbf{-1.2} \\ 
              3 & \checkmark & \checkmark & - & - & 72.5 & 81.9 & 70.9 & 67.1 & 72.9 & 70.0 & 91.6 & \textbf{+3.0}  \\ 
              4 & \checkmark & \checkmark & - & \checkmark & 72.9 & 82.3 & 71.0 & 66.8 & 72.7 & 70.0 & 91.9 & \textbf{+3.1} \\ 
              5 & \checkmark & - & \checkmark & - & 74.6 & 84.1 & 72.9 & 66.4  & 72.1 & 70.5 & 92.3 & \textbf{+3.9} \\ 
              6 & \checkmark & - & \checkmark & \checkmark & \textbf{75.2} & \textbf{84.8} & \textbf{73.4} & \textbf{67.4}  & \textbf{73.1} & \textbf{71.1} & \textbf{93.2} & \textbf{+4.6} \\ 
              \bottomrule
          \end{tabular}
          \caption{Ablation studies on RTrack.  $\Delta$ denotes the performance
              change (averaged over benchmarks). The superscript $\ddagger$ denotes that the tracker adopts a \textbf{one-to-one} assigner strategy, while all other assigners utilize an \textbf{one-to-many} strategy.}
          \label{Tab::ablation_all}
      \end{subtable}
%	}
  \begin{subtable}[t]{0.3\linewidth}
      \scriptsize
      %		\captionsetup{width=.95\linewidth}
%	\resizebox{\textwidth}{!}{
%   		\setlength\tabcolsep{3pt}
      \centering
      \begin{tabular}{c||ccc}
          \toprule
          %				\rowcolor{mygray}
          & ~ & ~ & ~ \\
          %				\rowcolor{mygray} 
          \multirow{-2}{*}{CD}
          & \multirow{-2}{*}{\makecell[c]{GOT-10k\\mAO(\%)$\uparrow$}}
          & \multirow{-2}{*}{\makecell[c]{LASOT\\AUC(\%)$\uparrow$}}
          & \multirow{-2}{*}{\makecell[c]{TKNET\\AUC(\%)$\uparrow$}} \\
          \midrule
          9 & 74.1 & 67.6 & 83.4 \\ 
          12 & \textbf{75.0} & \textbf{67.9} & \textbf{83.6} \\
          14 & 74.6 & 67.5 & 83.5 \\
          16 & 74.4 & 67.6 & 83.5 \\
          20 & 74.4 & 67.4 & 83.6 \\ 
          %		\hline
          \bottomrule
      \end{tabular}
%	}
      \caption{Top-k Center Distance (CD) candidate samples count.}
      \label{Tab::CD_candidate_cnt}
  \end{subtable}
  \hspace{0.3em}
  \begin{subtable}[t]{0.3\linewidth}
      \scriptsize
      %		\captionsetup{width=.95\linewidth}
%	\resizebox{\textwidth}{!}{
%		\setlength\tabcolsep{3pt}
      \centering
      \begin{tabular}{c||ccc}
          %		\hline
          \toprule
          %				\rowcolor{mygray}
          &  &  &  \\
          %				\rowcolor{mygray} 
          \multirow{-2}{*}{IV}
          & \multirow{-2}{*}{\makecell[c]{GOT-10k\\mAO(\%)$\uparrow$}}
          & \multirow{-2}{*}{\makecell[c]{LASOT\\AUC(\%)$\uparrow$}}
          & \multirow{-2}{*}{\makecell[c]{TKNET\\AUC(\%)$\uparrow$}} \\
          \midrule
          9 & 74.2 & 65.0 & 83.5 \\ 
          12 & 74.5 & 66.3 & \textbf{83.7} \\
          14 & 73.3 & 66.5 & \textbf{83.7} \\
          16 & 74.9 & 67.3 & 83.6 \\
          20 & \textbf{74.0} & \textbf{67.6} & 83.5 \\ 
          %		\hline
          \bottomrule
      \end{tabular}
%	}
      \caption{Top-k IoU Value (IV) candidate samples count.}
      \label{Tab::IV_candidate_cnt}
  \end{subtable}
  \hspace{0.3em}
  \begin{subtable}[t]{0.35\linewidth}
      %		\captionsetup{width=.95\linewidth}
%	\resizebox{\linewidth}{!}{
%		\setlength\tabcolsep{4pt}
      \scriptsize
      \begin{tabular}{cc|c||ccc}
          \toprule
          ~ & ~ & ~ & ~ & ~ &  \\
          \multicolumn{2}{c|}{\multirow{-2}{*}{\makecell[c]{Assigner}}} & \multirow{-2}{*}{\makecell[c]{w/\\leading}} &
          \multirow{-2}{*}{\makecell[c]{GOT\\-10k}} & \multirow{-2}{*}{\makecell[c]{LASOT}} & \multirow{-2}{*}{\makecell[c]{TKNET}} \\
          \hline
          \multicolumn{2}{c|}{\multirow{2}{*}{\makecell[c]{MaxIoU}}} & - & 72.1 & 65.8 & 82.3  \\
          ~ & ~ & $\checkmark$ & \textbf{72.9} & \textbf{66.8} & \textbf{83.4} \\
          \hline
          \multicolumn{2}{c|}{\multirow{2}{*}{\makecell[c]{CD\\(ours)}}} & - & 73.2 & 66.0 & 83.0  \\
          ~ & ~ & $\checkmark$ & \textbf{75.0} & \textbf{67.9} & \textbf{83.6} \\
          \hline
          \multicolumn{2}{c|}{\multirow{2}{*}{\makecell[c]{IV\\(ours)}}} & - & 72.0 & 64.9 & 82.2  \\
          ~ & ~ & $\checkmark$ & \textbf{75.2} & \textbf{67.4} & \textbf{83.5} \\ 
          %			\hline
          \bottomrule
      \end{tabular}
%	}
      \caption{Experiments on the impact of leading strategy.}
      \label{Tab::leading}
  \end{subtable}
  \caption{A set of ablative studies on GOT-10k, LASOT, OTB100 and TrackingNet(TKNET).}
\end{table*}

\section{Experiments}

\subsection{Implementation Details}
Our tracker is implemented using PyTorch. The models are trained on \textbf{one} NVIDIA RTX 3090 GPU with 24GB of memory each, and inference is performed on an NVIDIA RTX3060Ti GPU.

\subsubsection{Architectures.} 
The vanilla ViT-Base~\cite{MAE} model pre-trained with MAE~\cite{MAE} is adopted
as the backbone for our encoder. It contains 12 layers with a hidden dimension of 768.
We adopt an alternating training strategy, given RTrack with a total of 12 layers. In the shallow layers $l_0$ to $l_2$, we learn the importance of each patch token. The fine-tuned layers $(l_3, l_6, l_9)$ are trained on alternate epochs with $N$ fully dense patch tokens without discarding and $N'(\leq N)$ sparse patch tokens after discarding. Training with full dense patch tokens preserves the accuracy of the model, unlike dynamicViT~\cite{dynamicVit} and OSTrack~\cite{OSTrack}, which are unable to recover the original accuracy with dense tokens. We present the following RTrack versions that demonstrate its performance:

-RTrack-256. Template: 128~\texttimes~128 pixels; Search region: 256~\texttimes~256 pixels.

-RTrack-384. Template: 192~\texttimes~192 pixels; Search region: 384~\texttimes~384 pixels.

\subsubsection{Training.}
We use the LaSOT~\cite{LaSOT}, GOT-10k~\cite{GOT-10k}, COCO~\cite{COCO}, and TrackingNet~\cite{TrackingNet} datasets to train our RTrack.
We utilize the same data augmentations as
OSTrack~\cite{OSTrack}, including horizontal flip and brightness jittering.
Each GPU processes 64 pairs of images, resulting in a total batch size of 64. And we use the AdamW optimizer~\cite{adamw} with a weight decay of $10^{-4}$. The initial learning rate is $1.5 \times 10^{-4}$, which is reduced to $1.5 \times 10^{-5}$ at the 80\% of the total epochs. The training process of RTrack-256 includes 60k image pairs sampled for training per epoch.

\subsection{Comparison with State-of-the-art Trackers}
\label{Sec::sota}
To evaluate the performance of our proposed RTrack, we compare it with several state-of-the-art (SOTA) trackers on GOT-10k~\cite{GOT-10k} and TrackingNet~\cite{TrackingNet}. 
In addition to the mentioned datasets, the detailed results and analysis on UAV123~\cite{uav}, LaSOT~\cite{LaSOT}, and OTB-100~\cite{otb2015} can be found in the \textbf{appendix} of our paper.

\subsubsection{GOT-10k.}
The GOT-10k dataset is a comprehensive benchmark for object tracking. It comprises 10,000 video clips and approximately 1.5 million target annotations. The dataset covers 560 common object classes and 87 motion modes, making it diverse and representative of real-world tracking scenarios.
One notable feature of the GOT-10k dataset is the introduction of the one-shot protocol for training and testing. This protocol ensures that the training and test sets do not overlap in terms of object categories, allowing for better evaluation and assessment of the performance of non-specific target tracking tasks.
~\cref{Tab::sota_got10k} shows that RTrack-256 achieves the highest average overlap (AO) of 74.6\%, surpassing OSTrack-256 (71.0\%)~\cite{OSTrack} by 3.6\% and outperforming Swin-V2 (70.8\%)~\cite{swinv2}. It is worth noting that RTrack achieves these results with only $\frac{1}{3}$ of the training sample pairs used by OSTrack and $\frac{1}{18}$ of the training sample pairs used by Swin-V2. This demonstrates the effectiveness and efficiency of RTrack in extracting discriminative features for unseen classes.

\subsubsection{TrackingNet.}
The TrackingNet~\cite{TrackingNet} dataset is currently the largest and longest video object tracking dataset available. It consists of over 30,000 video sequences, encompassing various challenging scenarios, and includes more than 14 million object bounding box annotations.
~\cref{Tab::sota_tknet} shows that RTrack achieves superior performance compared to most trackers on the TrackingNet dataset. These results highlight the effectiveness of our proposed method, especially considering the smaller number of training samples used in our approach.

\subsection{Ablation and Analysis}
\label{Sec::abalation_study}
To verify the effectiveness of our proposed framework, we analyze different components and settings of RTrack and perform detailed exploration studies.
The result of the baseline is reported in ~\cref{Tab::ablation_all} (\#1). All Experiments are reported in ~\cref{Tab::ablation_all}, ~\cref{Tab::CD_candidate_cnt}, ~\cref{Tab::IV_candidate_cnt} and ~\cref{Tab::leading}.
More ablation studies and analyses are shown in the \textbf{appendix}.

%\begin{figurze}[!t]
%	\centering
%	\includegraphics[width = \linewidth]{figures/iou_curve.pdf}
%%	\caption{\textbf{Training IoU \textit{vs.} epoch.} The ``\textcolor[RGB]{196,49,25}{---}'' denote our proposed RTrack (w/ one-to-many assigner strategy), and ``\textcolor[RGB]{31,119,180}{---}'' denote the current \textbf{one-to-one} training efficiency tracker, OSTrack~\cite{}. Best viewed in color and zoom in.}
%	\label{fig::train_iou}
%\end{figure}

\subsubsection{Analysis on one-to-many assigner strategy in the refine stage.}
The design of the one-to-many assigner strategy in the refine stage allows for faster convergence speed with high performance. We also experimented with a commonly used static one-to-many sample assignment strategy, referred to as the ``MaxIoU assigner\footnote{The bounding box with an IoU (between the initial induced pseudo box and the ground-truth bounding box) larger than 0.5 is considered a positive sample, smaller than 0.4 is considered a negative sample, and any other IoU values are ignored.}."
As shown in ~\cref{Tab::ablation_all} (\#1, \#3, \#5), by comparing the performance of the one-to-one assigner strategy with the one-to-many strategy, we observed that the one-to-many strategy leads to improved higher performance. 
These results highlight the efficiency of our approach. By utilizing the one-to-many assigner strategy and optimizing the training process, we achieve comparable or even superior performance to existing state-of-the-art trackers while significantly reducing the training time. 

\subsubsection{Analysis on the correlation between classification and localization.}
Our baseline only considers the one-way information flow from classification (\textbf{cls}) to localization (\textbf{loc}).
To explore the impact of the mutual affinity between loc and cls,
we implicitly incorporate correlation modeling into classification and regression. This integration allows us to incorporate IoU information into the classification branch, thereby improving object localization.
By considering the correlation between cls and loc, i.e., \cref{Tab::ablation_all} (\#2, \#4, \#6), we enhance the integration of classification and localization information, leading to more accurate and reliable object tracking.

\subsubsection{Further analysis on the leading assigner strategy.}
The design of the ``leading'' is the key component of our RTrack in the refine stage. By enabling the refine stage to directly learn the information acquired by the init stage, the init stage can focus more on acquiring new information.
~\cref{Tab::leading} demonstrates the performance of our RTrack with the leading strategy and different assigner strategies. It shows that the leading strategy unleashes the potential of RTrack and improves its performance across different assigner strategies.
In the one-to-many assigner strategy, we also explored two sorting approaches based on Center Distance (CD) and IoU Value (IV), respectively, to obtain the top-k candidate samples. \cref{Tab::CD_candidate_cnt} and \cref{Tab::IV_candidate_cnt} show the settings under different ``k''.
This further demonstrates the good capacity of the design of the leading assigner strategy.

\section{Conclusion}
In this paper, we propose a baseline tracker to handle arbitrary representations, adaptively converting the sample points into pseudo bounding boxes in visual object tracking.
Additionally, we enhance the baseline tracker by introducing a one-to-many leading assignment strategy.
To the best of our knowledge, we are the first to  conduct an in-depth exploration of the training potential across multiple stages.
As pioneers in this approach, we present RTrack which significantly improves training convergence speed.

%%%%%%%%% REFERENCES
{\small
\bibliographystyle{ieee_fullname}
\bibliography{egbib}
}

\clearpage
\counterwithin{figure}{section}
\counterwithin{table}{section}

\appendix   %仅一个附录时用appendix，否则\appendices
\setcounter{table}{0}   %从0开始编号，显示出来表会A1开始编号
\setcounter{figure}{0}
\setcounter{equation}{0}
%定义编号格式，在数字序号前加字符“A"
\renewcommand{\thetable}{A\arabic{table}}
\renewcommand{\thefigure}{A\arabic{figure}}
\renewcommand{\theequation}{\arabic{equation}}
\section*{Supplementary Material}

\subsection*{A. Review of RepPoints}
\label{sec::rep_review}
RepPoints adopts pure regression to achieve object localization. Starting from a feature map position $p=(x,y)$, it directly regresses a set of points $\mathcal{S}' = \left\{p'_i = (x'_i, y'_i)\right\}_{i=1}^{n}$ to represent appearance using two progressive steps:
\begin{equation}
\begin{array}{c}
\mathbf{Step1:} p_{i} = p+\Delta p_{i} 
\\ = p+g_{i}\left(F_{p}\right) \\
\mathbf{Step2:} p_{i}^{\prime} = p_{i}+\Delta p_{i}^{\prime} \\ = p_{i}+g_{i}^{\prime}\left(\operatorname{concat}\left(\left\{F_{\mathbf{p}_{i}}\right\}_{i = 1}^{n}\right)\right),
\end{array}
\nonumber
\end{equation}
where $\mathcal{S} = \left\{p_i = (x_i, y_i)\right\}_{i=1}^{n}$ is the intermediate point set representation; $F_\mathbf{p}$ denotes the feature vector at position $p$; $g_i$ and $g'_i$ are 2-d regression functions implemented by a linear layer. The pseudo bounding box is obtained by applying a conversion function $\mathcal{T}$ on the point sets $\mathcal{S}$ and $\mathcal{S}'$, where $\mathcal{T}$ is modeled as the \emph{min-max} or \emph{moment} function.

To achieve accurate SOT, a set of adaptive sample points are modeled in our proposed method, which we refer to as \textit{TrackPoints}. The sample points are defined as
\begin{equation}
\mathcal{S} = {(x_k, y_k)}_{k=1}^{n},
\end{equation}
where $n$ is the number of sample points used in the representation. In our approach, we set $n$ to be 9 by default.

To improve the localization accuracy of the tracked object, we progressively refine the position of the \textit{TrackPoints}. The refinement process can be expressed as
\begin{equation}
\label{eq::dbox_refine_sot}
\mathcal{S}r = {(x_k + \Delta x_k, y_k + \Delta y_k)}_{k=1}^{n},
\end{equation}
where ${(\Delta x_k, \Delta y_k)}_{k=1}^{n}$ are the predicted offsets of the new sample points with respect to the old ones. The refinement process implicitly considers the changes in the object's position and appearance.

To evaluate the performance of our proposed SOT method, we need to convert the \textit{TrackPoints} into a bounding box for comparison with the ground-truth bounding box. We use a predefined converting function $\mathcal{T}: \mathcal{T}_O \rightarrow \mathcal{B}_O$, where $\mathcal{T}_O$ denotes the \textit{TrackPoints} for object $O$ and $\mathcal{B}_O$ represents a \textit{pseudo box}.

Three converting functions are considered in our approach:
\begin{itemize}
    \setlength{\itemsep}{0pt}
    \setlength{\parsep}{0pt}
    \setlength{\parskip}{0pt}
    \item \textbf{Min-max function.} This function performs a min-max operation over both axes of the \textit{TrackPoints} to determine $\mathcal{B}_O$, which is equivalent to the bounding box over the sample points.
    \item \textbf{Moment-based function.} This function computes the center point and scale of the rectangular box $\mathcal{B}_O$ using the mean value and the standard deviation of the \textit{TrackPoints}, where the scale is multiplied by globally-shared learnable multipliers $\lambda_x$ and $\lambda_y$.
\end{itemize}
All three converting functions are differentiable, which allows us to perform end-to-end learning when they are incorporated into the SOT system.

\subsection*{B. Differences with Reppoints}

Our proposed RTrack shares a similar spirit with Reppoints~\cite{reppointv1}. Both of them cast the object representation as a set of sample points.
%and automatically arrange the points to define the spatial extent of objects. 
However, our method differs from Reppoints in three fundamental ways. 1) The tasks are different, Reppoints is designed for the object detection (OD) task, while ours is for tracking. 2)  The architectures are different, Reppoints adopts ResNet~\cite{ResNet_2016_CVPR} as its backbone network followed by  feature pyramidal networks (FPN)~\cite{FPN_2017_CVPR}.
Our method is more compact, only using a single encoder transformer without hierarchy features. RTrack employs ViT~\cite{vit} as the encoder for feature extraction. 3) The sample allocation strategies are different. For the refine stage, Reppoints selects positive and negative samples based on a hand-crafted static IoU threshold. Single object tracking (SOT) merely involves a single target, the choice of strategy is crucial for training stability and convergence speed. In contrast, our sample allocation strategy utilizes the one-to-many k-center nearest neighbors (KCNN) to obtain \textbf{candidate} positive samples. And further, we further divide positive and negative samples for effective training based on the dynamic threshold derived from mean and variance. 4) The perspective on classification and regression is different. In Reppoints, the classification task typically involves a multi-class detection problem, eg. 80 classes in the COCO~\cite{COCO} dataset. One-hot labels and a cross-entropy loss naturally match the requirements of multi-class classification. In our task, we treat classification as a foreground-background segmentation problem. After truncating the gradient information of the regression branch, we potentially combine classification and IoU collaboratively to enhance performance.

\subsection*{C. Performance metrics}
The official online evaluation server for the GOT-10k~\cite{GOT-10k} dataset uses the average overlap (AO), success rate ($SR_{0.5}$), and success rate ($SR_{0.75}$) as evaluation criteria.
\begin{itemize}
    \item Average Overlap (AO): This measures the average intersection-over-union (IOU) between the predicted bounding box and the ground truth over the entire sequence. AO is calculated as follows:
    \begin{equation}
    A O=\frac{1}{N} \sum_{i=1}^N I O U_i
    \end{equation}
    \item Success Rate ($SR_{0.5}$): This measures the proportion of frames for which the IOU between the predicted bounding box and the ground truth is greater than or equal to 0.5. $SR_{0.5}$ is calculated as follows:
    \begin{equation}
    S R_{0.5}=\frac{1}{N} \sum_{i=1}^N\left[I O U_i \geq 0.5\right]
    \end{equation}
    where $[IOU_i \geq 0.5]$ equals 1 if $IOU_i \geq 0.5$.
    \item $SR_{0.75}$ is similar to $SR_{0.5}$, but it measures the success rate at a higher overlap threshold of 0.75.
\end{itemize}

\begin{table*}[!ht]
    \centering
    %	\resizebox{\linewidth}{!}{
    \begin{tabular}{c||cccccccccc||c}
        \toprule
        ~ & \multirowcell{2}[0pt][c]{Siam\\FC++} & 
        \multirowcell{2}[0pt][c]{DiMP} & 
        \multirowcell{2}[0pt][c]{Pr\\DiMP} & 
        \multirowcell{2}[0pt][c]{Tr\\DiMP} & 
        \multirowcell{2}[0pt][c]{Siam\\RCNN} & 
        \multirowcell{2}[0pt][c]{TransT} & 
        \multirowcell{2}[0pt][c]{STARK\\ST101} & 
        \multirowcell{2}[0pt][c]{ToMP\\50} & 
        \multirowcell{2}[0pt][c]{MixFormer\\1k} &
        \multirowcell{2}[0pt][c]{MAT} & 
        \multirowcell{2}[0pt][c]{RTrack\\(ours)} \\
        ~ & ~ & ~ & ~ & ~ & ~ & ~ & ~ & ~ & ~ & ~ & ~\\

        % ~ & \multirowcell{3}[0pt][c]{Siam\\FC++\\~\cite{siamFC++}} & 
        % \multirowcell{3}[0pt][c]{DiMP\\~\cite{DiMP}} & 
        % \multirowcell{3}[0pt][c]{Pr\\DiMP\\~\cite{prdimp}} & 
        % \multirowcell{3}[0pt][c]{Tr\\DiMP\\~\cite{trdimp}} & 
        % \multirowcell{3}[0pt][c]{Siam\\RCNN\\~\cite{SiamR-CNN}} & 
        % \multirowcell{3}[0pt][c]{TransT\\~\cite{TransT}} & 
        % \multirowcell{3}[0pt][c]{STARK\\ST101\\~\cite{STARK}} & 
        % \multirowcell{3}[0pt][c]{ToMP\\50\\~\cite{tomp}} & 
        % \multirowcell{3}[0pt][c]{MixFormer\\1k\\~\cite{Mixformer}} &
        % \multirowcell{3}[0pt][c]{MAT\\~\cite{mat}} & 
        % \multirowcell{3}[0pt][c]{RTrack\\(ours)} \\
        % ~ & ~ & ~ & ~ & ~ & ~ & ~ & ~ & ~ & ~ & ~ & ~\\

        \midrule
        UAV123 & - & 65.4 & 68.0 & 67.5 & 64.9 & 69.1 & 68.2 & 69.0 & 68.7 & 68.0 & \textbf{69.5} \\ 
        LASOT & 54.4 & 56.9 & 59.8 & 63.9 & 64.8 & 64.9 & 67.1 & 67.6 & 67.9 & 65.6 & \textbf{68.9} \\
        OTB100 & 68.3 & 68.4 & 69.6 & 71.1 & 70.1 & 69.4 & 68.1 & 70.1 & 70.4 & - & \textbf{71.1} \\ 
        \bottomrule
    \end{tabular}
    %	}
    \caption{Comparision with state-of-the-art trackers on the LASOT~\cite{LaSOT}, UAV123~\cite{uav}, OTB100~\cite{otb2015} datasets in terms of AUC(\%).}
    \label{Tab::sota_lasot_uav_otb}
\end{table*}

We use $AUC$, $P_{norm}$, and $P$ as evaluation metrics for all datasets except the GOT-10K dataset.
AUC (Area Under the Curve) measures the area under the success plot, which is the curve of success rate versus overlap threshold. $P_{norm}$ is the normalized precision at 20 pixels, which measures the precision of the predicted bounding box normalized by the size of the ground truth box. $P$ is the precision score, which measures the percentage of frames where the predicted box and ground truth box have an overlap greater than a certain threshold.

\begin{figure}[!t]
    \begin{minipage}[t]{\columnwidth}
        \centering
        \includegraphics[width=\textwidth]{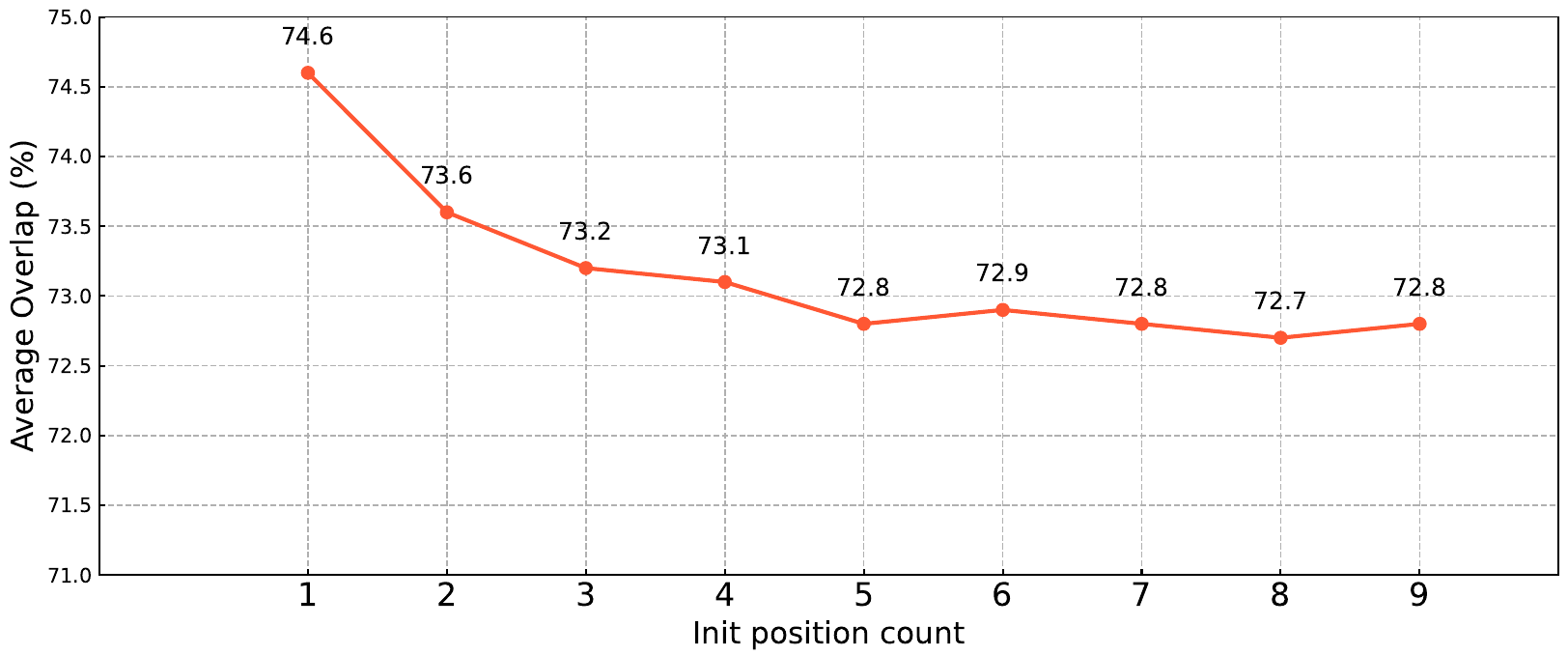}
    \end{minipage}
    \begin{minipage}[t]{\columnwidth}
        \centering
        \includegraphics[width=\textwidth]{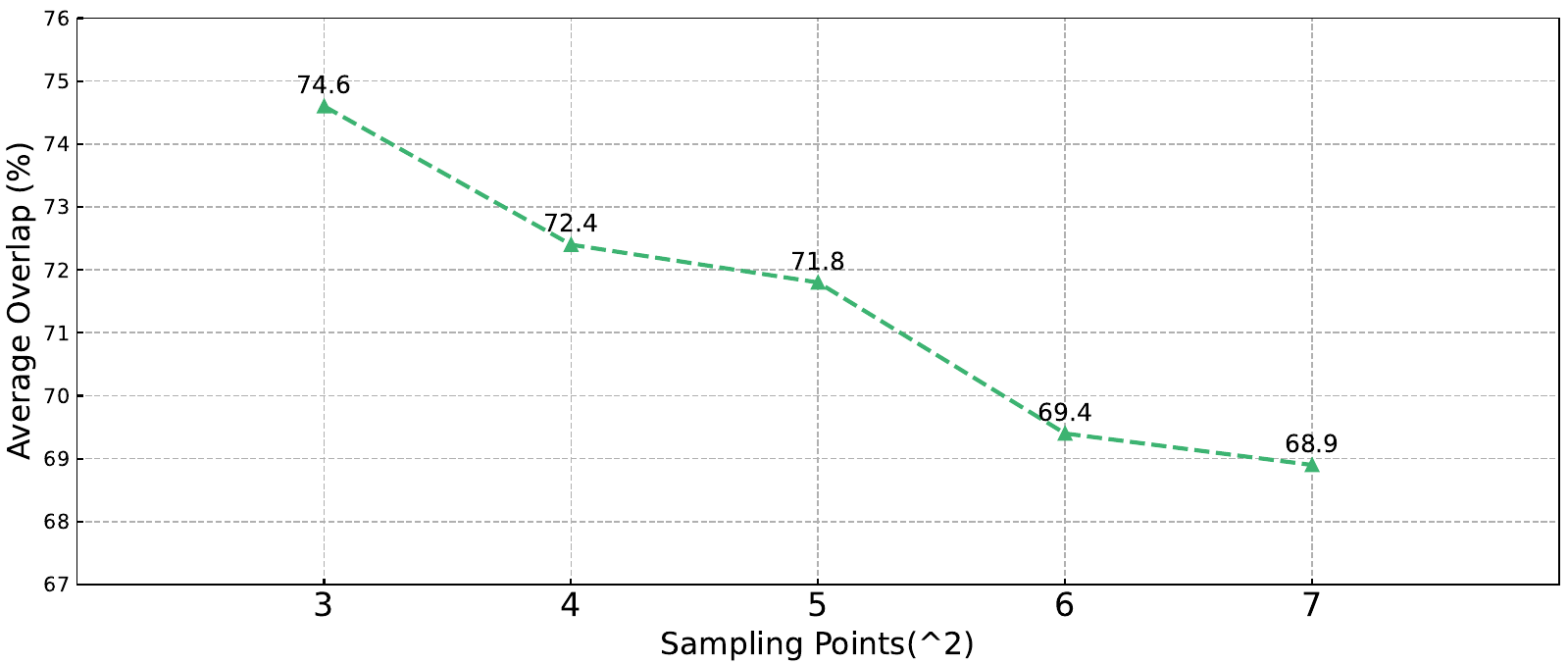}
    \end{minipage}
    \caption{Influence of the init position count and sample points on GOT-10k~\cite{GOT-10k}.}
    \label{fig::basic_settings}
\end{figure}

\subsection*{D. Experiments on UAV123, LASOT and OTB-100}
UAV123\cite{uav} is a dataset comprising 123 video sequences captured from the perspective of a low-altitude unmanned aerial vehicle. With an average sequence length of 915 frames, it is designed for long-term tracking evaluation.
LaSOT~\cite{LaSOT} is also a benchmark specifically designed for long-term tracking evaluation. It features a test set comprising 280 videos and a total of 1,400 video sequences covering 70 different classes of objects.
The Object Tracking Benchmark (OTB)~\cite{otb2015} is a benchmark that evaluates the performance of visual tracking algorithms.
The results in~\cref{Tab::sota_lasot_uav_otb} show that RTrack has superior performance compared to most trackers on all three benchmarks, demonstrating the
strong generalizability of RTrack.

\begin{table*}[t]
    \begin{subtable}[t]{0.35\linewidth}
        \begin{tabular}{c|cc}
            \toprule
            & ~  &  ~ \\
            \multirow{-2}{*}{\makecell[c]{~}}
            & \multirow{-2}{*}{\makecell[c]{GOT-10k\\mAO(\%)$\uparrow$}}
            & \multirow{-2}{*}{\makecell[c]{LASOT\\AUC(\%)$\uparrow$}} \\
            \midrule
            \multirowcell{2}[0pt][c]{w/o truncation\\ gradient} & \multirowcell{2}[0pt][c]{N/A} & \multirowcell{2}[0pt][c]{N/A} \\ \\ \midrule
            pos & 74.3 & 66.8\\
            pos \& neg & 73.7 & 66.4\\
            %		\hline
            \bottomrule
        \end{tabular}
        \caption{Correlation settings. 'N/A' means unstable inference.}
        \label{Tab::corr_pos_neg}
    \end{subtable}
    \hspace{2em}
    \begin{subtable}[t]{0.6\linewidth}
        \scriptsize
        \centering
        \resizebox{\textwidth}{!}{
            \begin{tabular}{c||ccc|cc|cc||c}
                \toprule
                ~ & \multicolumn{3}{c|}{GOT-10k} & \multicolumn{2}{c|}{LASOT}  & \multicolumn{2}{c||}{TKNET} & \multirowcell{2}[0pt][c]{Training \\Pairs\\($\times10^{6}$)} \\ 
                \multirow{-2}{*}{Trackers} & \makecell[c]{mAO \\ (\%)} & \makecell[c]{mSR$_{50}$ \\ (\%)} & \makecell[c]{mSR$_{75}$ \\ (\%)} & \makecell[c]{SUC \\ (\%)} & \makecell[c]{PRE \\ (\%)} & \makecell[c]{SUC \\ (\%)} & \makecell[c]{RPE \\ (\%)} & ~ \\ 
                \midrule
                RTrack$_{100}$ & 74.4 & 84.1 & 72.3 & 67.4  & 73.4 & 83.6 & 82.5 & 6  \\
                RTrack$_{300}$ & 75.9 & 85.2 & 74.8 & 67.0  & 72.7 & 84.2 & 83.3 & 18  \\ 
                RTrack$_{500}$ & 74.2 & 83.6 & 72.7 & 68.1  & 74.3 & 83.9 & 82.7 & 30  \\ 
                \bottomrule
            \end{tabular}
        }
        \caption{RTrack extension as epochs (training pairs) increasing. Note: for each tracker, we performed a full round of training (i.e. the decay epoch of the learning rate is always at 0.8 of the total epoch).}
        \label{Tab::epoch_extension}
    \end{subtable}
    \caption{More \textbf{ablative studies} on GOT-10k, LASOT and TrackingNet(TKNET).}
\end{table*}

\subsection*{E. More ablation studies and analyses}
\label{appendix::settings}
\begin{figure}[!t]
    \centering
    \includegraphics[width = \linewidth]{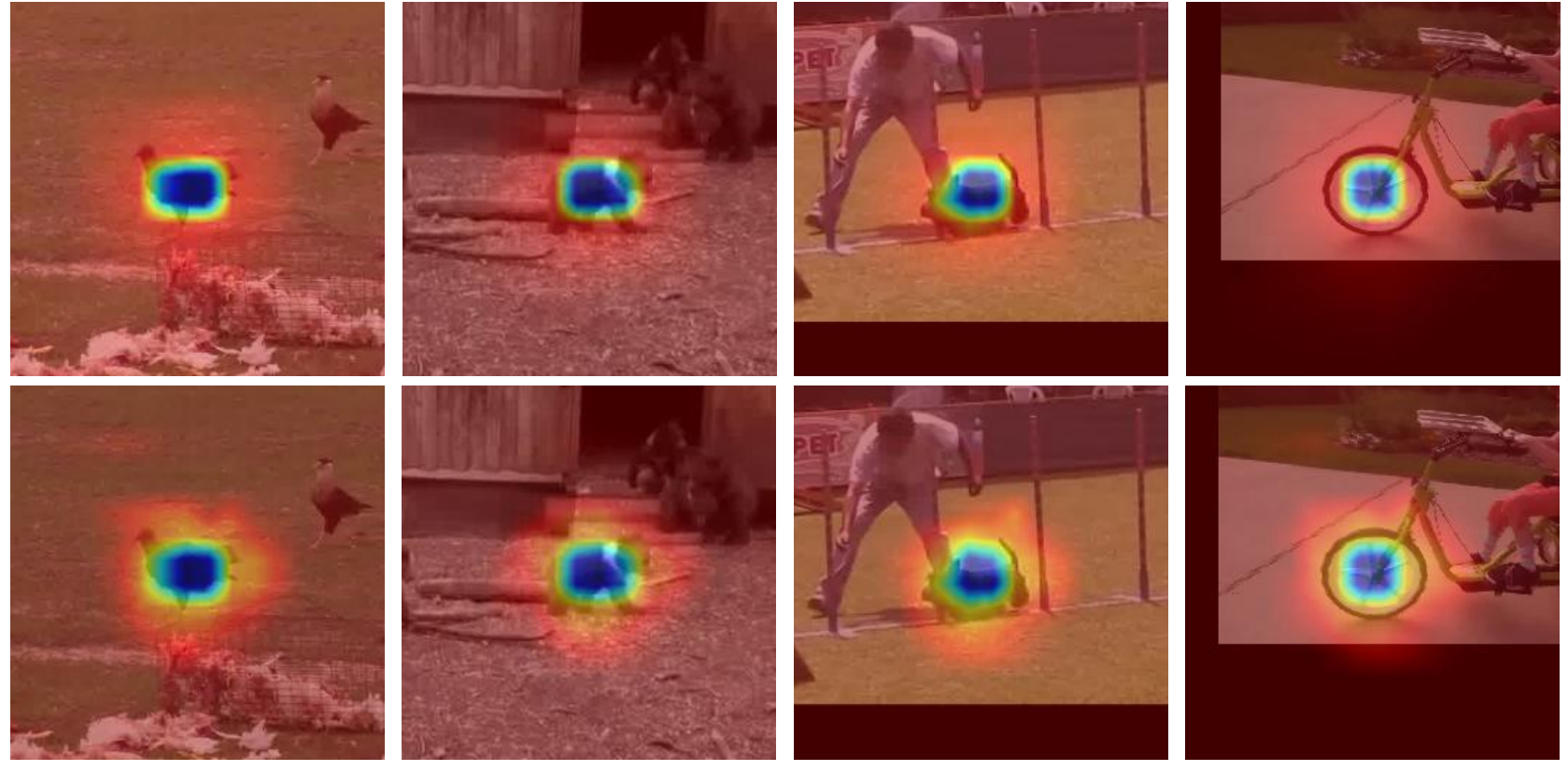}
    \caption{The distribution of classification scores in the search region.  In the top row, where there is no correlation between classification and localization. In the bottom row, the mutual affinity between classification and localization is modeled.}
    \label{fig::corr_heatmap}
\end{figure}

\noindent \textbf{Mutual affinity of pos \& neg samples.}
To investigate the impact of the correlation between classification and localization on positive and negative samples, we limited the correlation to positive samples only. In \cref{Tab::corr_pos_neg}, we extend the correlation to both positive and negative samples. We found that extending the correlation to both positive and negative samples did not have a significant impact on the results. In our design, we implicitly incorporate correlation modeling into classification and regression. This integration allows us to incorporate IoU information into the classification branch, thereby improving object localization, as shown in ~\cref{fig::corr_heatmap}.

\noindent \textbf{Init position count in baseline tracker.}
To investigate the impact of the number of initial point representations on performance, in our proposed baseline RTrack, we use center point-based representation as the initial point representation in the \textbf{init} stage. Therefore, in ~\cref{fig::basic_settings}, we explored this setting by expanding it to \{2, 3, 4, 5, 6, 7, 8, 9\}. We found that increasing the init position count degrades the tracking performance. This indicates that the center point-based representation is beneficial for the representation of the target and the convergence of training.

\noindent \textbf{Sampling points count in baseline tracker.}
To further explore whether increasing the number of sampling points is beneficial, we gradually increased the number of sampling points from 9 to \{16, 25, 36, 49\}. As shown in ~\cref{fig::basic_settings}, we found that increasing the number of sampling points does not lead to performance improvement. This suggests that simply increasing the number of sample points may degrade tracking performance, perhaps because objects may face complex background effects, such as occlusion, as the track progresses.

\noindent \textbf{Analysis on the extension of RTrack.}
To explore whether the increase in training samples can improve performance, we gradually increase the number of epochs to \{100, 300, 500\} to align with the training cost of current state-of-the-art trackers, in ~\cref{Tab::epoch_extension}. 
The refine stage adopts a sorting strategy based on Center distance (CD), where the top 20 samples are selected as candidate samples.
As a result, we found that the proportional increase in training time resulted in only a slight performance improvement on short-term tracking datasets.

\begin{figure}[!t]
    \centering
    \includegraphics[width = \linewidth]{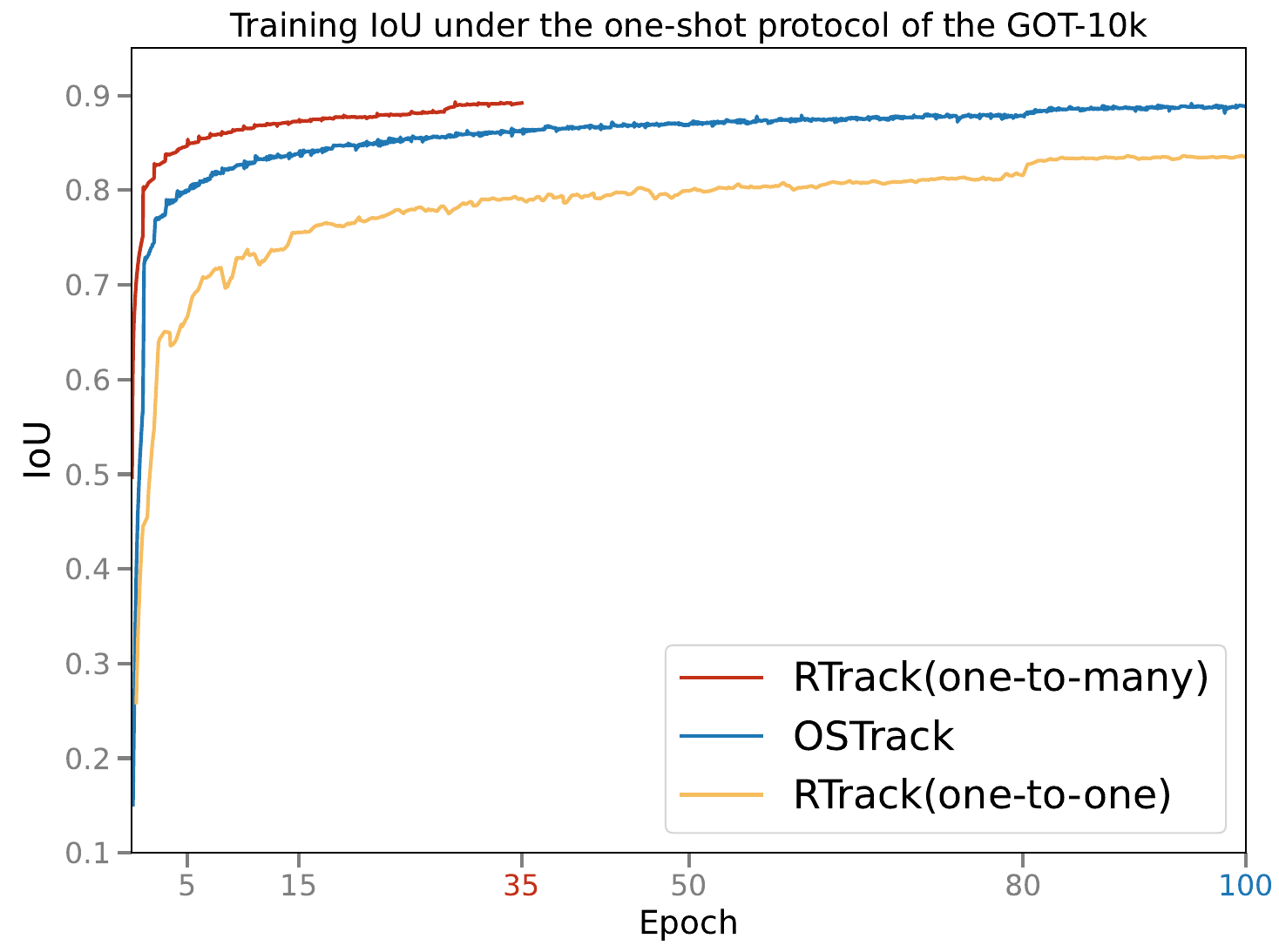}
    \caption{\textbf{Training IoU \textit{vs.} epoch.} Best viewed in color and zoom in.}
    \label{fig::train_iou_appendix}
\end{figure}

\noindent \textbf{Analysis on training convergence potential.}
To better illustrate the improvement in training convergence speed achieved by the one-to-many leading strategy, we compare our baseline tracker using the one-to-one matching strategy with our proposed one-to-many leading assignment strategy. As shown in ~\cref{fig::train_iou_appendix}, we observe that the simple one-to-one strategy, which only assigns a unique positive sample without incorporating tricks for localization, exhibits the slowest convergence speed. On the other hand, the one-to-one matching strategy used in training efficiency OSTrack~\cite{OSTrack}, which incorporates gaussian focal loss~\cite{focal}, further improves the training convergence speed. The experimental results demonstrate that our one-to-many leading assignment strategy can further tap into the training convergence potential of the tracker.

\begin{figure}[!t]
    \centering
    \includegraphics[width = \linewidth]{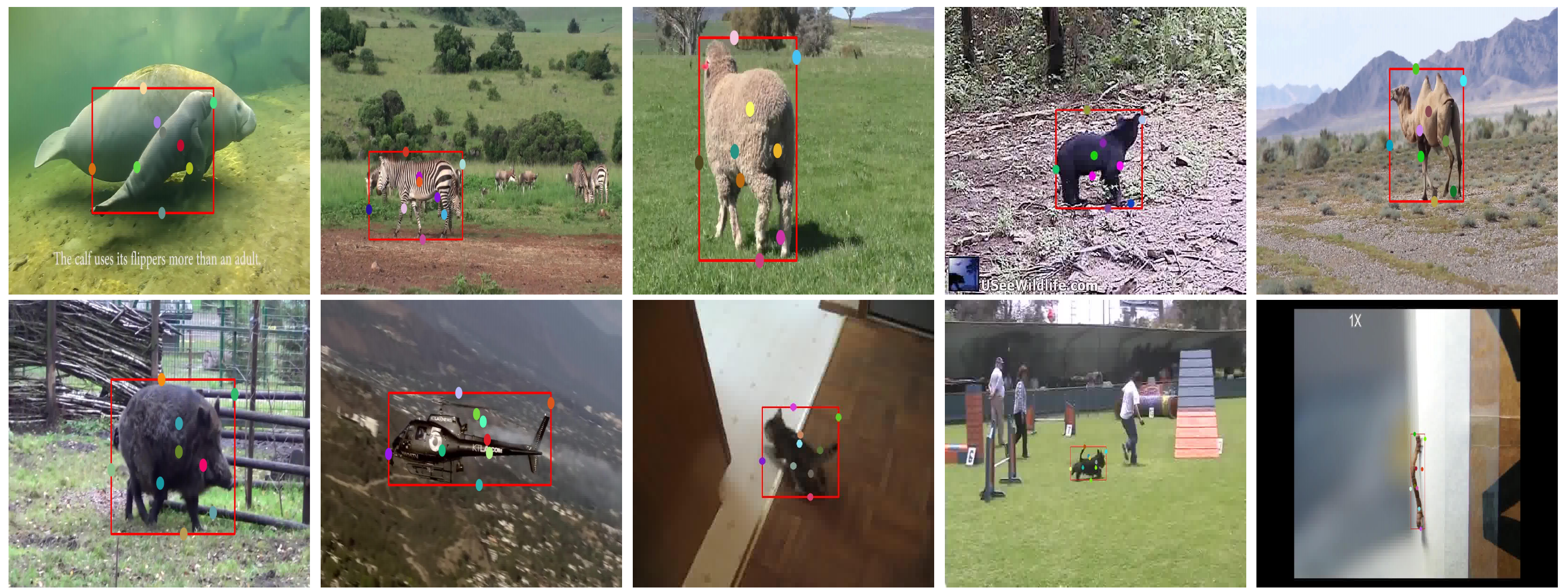}
    \caption{Point representation. We employ a min-max operation over both axes of the \textit{TrackPoints} to determine the final pseudo bounding box.}
    \label{fig::point_presentation}
\end{figure}

\noindent\textbf{Visualization on non-rigid representation.}
To gain a better understanding of our RTrack, we generate a plot of the sampling points while predicting points coordinate. To test the robustness of our model, we use complex scenarios encountered in real-world trackings, such as rotation, and background clutter, as shown in ~\cref{fig::point_presentation}. Interestingly, our tracker focuses on the appropriate extremities when predicting each coordinate, demonstrating our model’s ability for precise localization.

% \subsection*{F. Limitations}
% It is important to note that our algorithm relies on pre-trained weights for feature extraction, although it is common in current popular designs~\cite{OSTrack, GRM, Mixformer, seqtrack, artrack,simTrack}. Additionally, the encoder and the arbitrary representation head components remain independent in our current framework. In future work, we aim to merge the encoder and arbitrary representation head into a more tightly structure. This integration would provide a more unified and robust tracking solution.

\end{document}